\documentclass[10pt,twocolumn,letterpaper]{article}

\usepackage[pagenumbers]{cvpr} 

\newcommand{\red}[1]{{\color{red}#1}}

\definecolor{cvprblue}{rgb}{0.21,0.49,0.74}
\usepackage[pagebackref,breaklinks,colorlinks,allcolors=cvprblue]{hyperref}
\usepackage{xcolor}

\newcommand{\blue}[1]{{\color{blue}{[#1]}}}
\newcommand{\vsp}{\vspace{5pt}}
\renewcommand{\red}[1]{{\color{red}{[#1]}}}

\def\paperID{*****} 
\def\confName{CVPR}
\def\confYear{2026}

\title{Dual Diffusion Models for Multi-modal Guided 3D Avatar Generation}

\author{
    Hong Li$^{1}$\thanks{Corresponding author.} \quad 
    Yutang Feng$^{1}$ \quad 
    Minqi Meng$^{1}$ \quad 
    Yichen Yang$^{1}$ \quad 
    Xuhui Liu$^{2}$ \quad 
    Baochang Zhang$^{1}$ \\
    $^{1}$Beihang University \quad $^{2}$KAUST \\
    {\tt\small link0502@buaa.edu.cn} 
}

\begin{document}
\maketitle
\begin{abstract}
Generating high-fidelity 3D avatars from text or image prompts is highly sought after in virtual reality and human-computer interaction. However, existing text-driven methods often rely on iterative Score Distillation Sampling (SDS) or CLIP optimization, which struggle with fine-grained semantic control and suffer from excessively slow inference. Meanwhile, image-driven approaches are severely bottlenecked by the scarcity and high acquisition cost of high-quality 3D facial scans, limiting model generalization. To address these challenges, we first construct a novel, large-scale dataset comprising over 100,000 pairs across four modalities: fine-grained textual descriptions, in-the-wild face images, high-quality light-normalized texture UV maps, and 3D geometric shapes. Leveraging this comprehensive dataset, we propose PromptAvatar, a framework featuring dual diffusion models. Specifically, it integrates a Texture Diffusion Model (TDM) that supports flexible multi-condition guidance from text and/or image prompts, alongside a Geometry Diffusion Model (GDM) guided by text prompts. By learning the direct mapping from multi-modal prompts to 3D representations, PromptAvatar eliminates the need for time-consuming iterative optimization, successfully generating high-fidelity, shading-free 3D avatars in under 10 seconds. Extensive quantitative and qualitative experiments demonstrate that our method significantly outperforms existing state-of-the-art approaches in generation quality, fine-grained detail alignment, and computational efficiency.
\end{abstract}    
\section{Introduction}
\begin{figure*}[thb]
  \centering
  \includegraphics[width=1\linewidth]{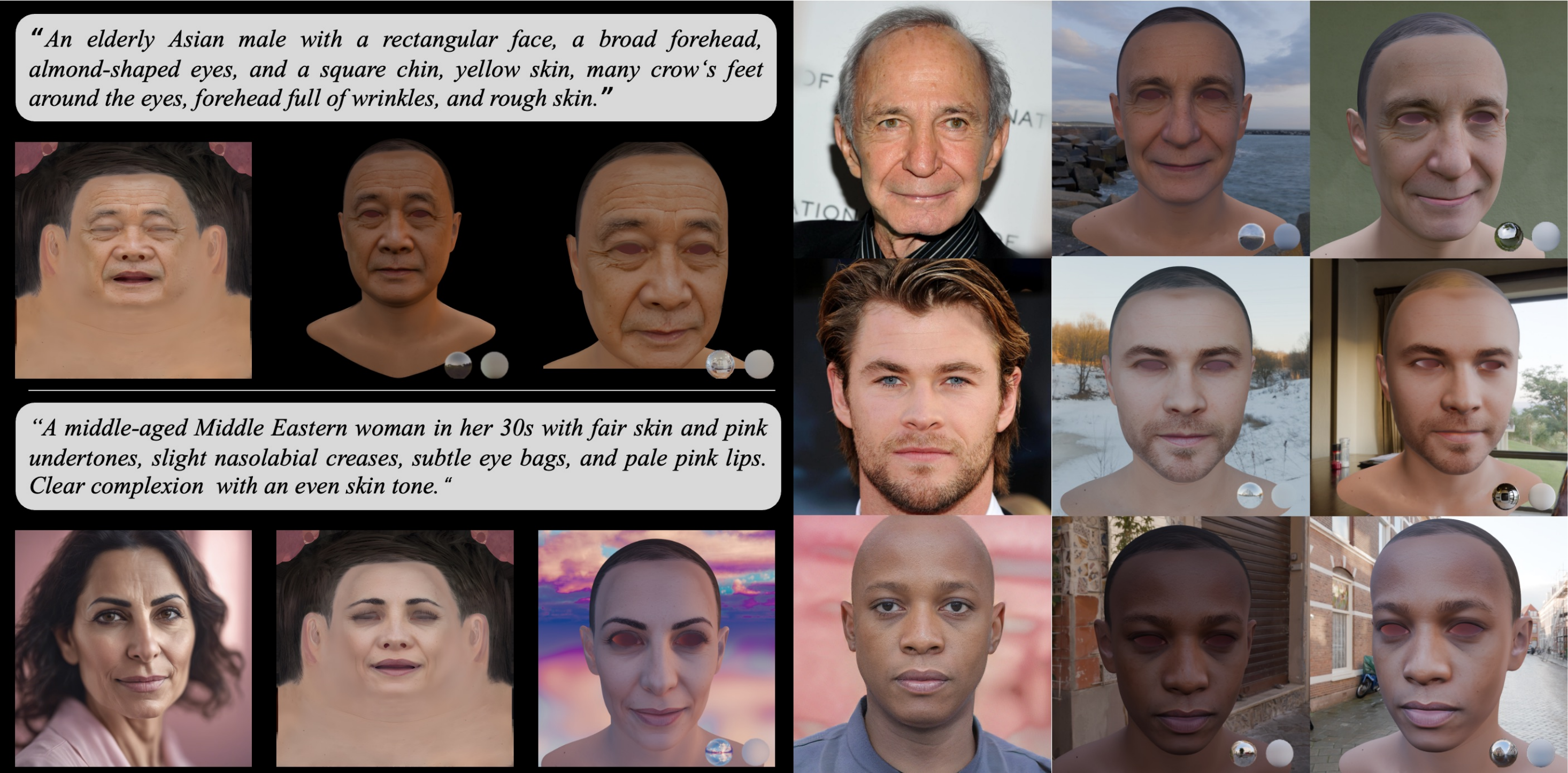}
  \caption{PromptAvatar generates realistic and animatable 3D avatars from a single text prompt, image, or both, compatible with 3D rendering engines like Blender. The top-left corner uses a text prompt to create an accurate texture UV-map and mesh. The bottom-left corner combines text with an FLUX-dev-1.0-generated~\cite{team2025zimage} image to guide high-quality texture UV-map creation. When an image prompt is used, facial geometry is extracted via a pre-trained 3D face reconstruction network \cite{bai2022ffhq, deep3d2020}. On the right, image prompts enable detailed texture effects like crow’s feet and beards.}
  \label{fig:intro}
\end{figure*}
Creating realistic 3D avatars based on text or image prompts holds significant potential in various fields such as virtual reality, augmented reality, education, artistic creation, gaming, film production, and human-computer interaction \cite{Lattas2023fitme, Papantoniou2023Relightify,zhou2024ultravatar, wu2023high}.
In the field of text-to-avatar, a recent wave of work has emerged, such as \cite{zhang2023dreamface, zhang2023text} using the SDS loss \cite{poole2022Dreamfusion, wang2023prolificdreamer} to constrain generated 3D faces to match given facial attribute descriptions. However, SDS loss tends to produce smooth results, sacrificing the diversity and controllability of text-to-3D face mapping and requiring long training for each text description. Moreover, methods based on the SDS loss usually require iterative fitting and are time-consuming. Some approaches \cite{rowan2023text2face, wu2023high} use CLIP \cite{mohammad2022clip} to find the relationship between text and 3D facial texture UV-map and shape from a facial image. However, due to being trained on generic data, CLIP has limited ability to distinguish facial details and often introduces uncontrollable lighting, making it difficult to use directly as digital assets. Both approaches struggle to support fine-grained text prompt generation. Intuitively, training a model from a large dataset containing pairs of detailed facial description text and high-quality 3D information (texture UV-map and shape) could significantly address these issues. However, such data is rare.

In the image-to-avatar domain, many efforts have been made to improve 3D geometric reconstruction accuracy \cite{deep3d2020, Feng2021deca, danvevcek2022emoca, Lei2023hrn, MICAECCV2022}, with only a few works \cite{gecer2021fast, lattas2020avatarme, Lattas2022AvatarMe++, Papantoniou2023Relightify, Lattas2023fitme, zhou2024ultravatar, bai2022ffhq} focusing on recovering UV texture maps from images. Unlike previous methods that estimate linear coefficients that struggle to express high-frequency facial texture \cite{deep3d2020, Feng2021deca}, \cite{lattas2020avatarme, Lattas2022AvatarMe++, Papantoniou2023Relightify, Lattas2023fitme} attempt to collect facial images and their 3D textures from the real world using specialized equipment to establish a mapping from images to 3D texture UV-map, and then train generative adversarial networks (GAN) \cite{karras2020analyzing}, diffusion models (DM) \cite{ho2020denoising}, or other generative models to recover textures from wild images. However, the cost of such data is high, the data amount is limited, and ensuring the model's generalization is challenging. FFHQ-UV \cite{bai2022ffhq} utilizes StyleGAN2's \cite{karras2020analyzing} powerful generation and attribute-editing capabilities \cite{Abdal2021styleflow} to create a large-scale high-quality normalized texture dataset for training generative models, but it lacks data that matches wild 2D facial images. Here, we define high-quality normalized facial texture UV-maps as those that are evenly illuminated and free from undesired occlusions, such that the texture UV-maps can be used as facial assets for rendering under different lighting conditions.

Benefiting from these works, in this paper, we first create a large-scale dataset based on synthetic data containing four modalities: fine-grained descriptions of facial attributes annotated by Qwen2.5-VL-32B-Instruct \cite{Qwen2.5-VL}, wild images with complex lighting and poses, light-normalized texture UV-maps, and 3D geometric identity coefficients corresponding to facial features. The data curation pipeline is cost-effective and highly reproducible. Furthermore, we propose \textbf{PromptAvatar}, consisting of a texture diffusion model and a geometric diffusion model. The former allows us to generate accurate and realistic high-resolution facial textures from text descriptions or image prompts. The latter can generate geometric structures from a single text prompt. Based on the large-scale dataset and the diffusion models with powerful conditional generation capabilities, we do not need to use the SDS loss or CLIP as a bridge for mapping text or 2D images to the 3D space. Instead, we directly train the model to generate 3D avatars guided by text or image prompts.
As shown in Figure \ref{fig:intro}, we demonstrate the generation results and the corresponding relighting using the Blender rendering engine.

Our key contributions are summarized as follows.
\begin{itemize}
\item
We construct a low-cost and reproducible large-scale dataset consisting of over a hundred thousand pairs of fine-grained text descriptions and 3D facial information.
\item
We introduce PromptAvatar, with texture and geometry diffusion models, enabling the generation of high-quality 3D avatars based on text and/or image prompts within 10 seconds.
\item
Through extensive qualitative and quantitative experiments, we demonstrate the superior performance of our dataset and model.
\end{itemize}
\section{Related Work}
\textbf{Portrait Relighting} Portrait relighting involves recovering a face image’s albedo and lighting information from a single image and re-rendering it under different lighting. 2D deep learning methods~\cite{Hou2021Towards, Andrew2022Face, Nestmeyer_2020_CVPR, Pandey2021Total, Sengupta2022SfSNet, Sun2019Single, Zhou2019DPR} typically require explicit supervision with multiple images of a single identity captured under varying lighting conditions. ~\cite{Yeh2022Learning} introduces a virtual light stage using advanced computer graphics to bridge the gap between synthetic and real data. The rise of 3D perception GANs \cite{Chan2022EG3D} reduces the need for estimating 3D information from 2D images for relighting. ~\cite{neural-3D-relightable} integrates the Phong albedo model into a neural volumetric rendering framework, enabling the learning of facial shape and material properties from in-the-wild 2D images without manual annotations, achieving controllable relighting. LumiGAN ~\cite{deng2023lumigan} uses self-supervised learning to relate the visibility equation to 3D shape, generating surface normals, diffuse albedos, and specular highlights for relighted faces. Our approach to light-normalized face image generation is inspired by NeRFFaceLighting \cite{Jiang2023NeRFFaceLighting}, which generates 3D-aware face images with reasonable albedo and shadow components using an implicit lighting representation.

\vsp

\textbf{Image-to-Avatar Generation} The generation of 3D faces has seen rapid progress in recent years. Initially, avatar generation heavily depends on intricate and expensive scanning setups, constraining its scalability \cite{alexander2010digital, borshukov2005realistic, guo2019relightables, lombardi2018deep}. Later research efforts attempt to generate realistic facial modeling, thereby establishing a solid groundwork for further research in image-to-avatar generation. Based on Principal Component Analysis (PCA), parametric models are used to express facial components containing identity, expression, and texture \cite{cao2013facewarehouse, li2017learning, ranjan2018generating}. Using differentiable rendering techniques, various methods employ self-supervised or weakly supervised learning to achieve high-fidelity avatar reconstruction. For example, \cite{deep3d2020, Feng2021deca} use mixed-level image information to achieve faithful face reconstruction and obtain more realistic facial textures. Some methods \cite{Tran_Liu_2018, tran2019towards, saito2017photorealistic} use 2D UV texture representations, which help neural networks obtain high-quality rendered images. The texture decoder-based methods \cite{Tran_Liu_2018, tran2019towards, Ron2022Unsupervised, Gecer2021OSTeC, bai2022ffhq, Lattas_2023_CVPR} take advantage of StyleGAN2's \cite{karras2020analyzing} ability to generate high-resolution UV images. Then, the 3D Morphable Model (3DMM) matching algorithm is used to find the best latent code for facial mesh generation. The iterative strategy from coarse to fine is used by both HRN \cite{Lei2023hrn} and NextFace \cite{Dib2021Practical} to get realistic results when reconstructing geometric details and textures. Nevertheless, these iterative fitting approaches are susceptible to overfitting when dealing with occlusions like glasses or hairs. Relightify \cite{Papantoniou2023Relightify} utilizes a diffusion model \cite{rombach2022high}, and FitMe \cite{Lattas_2023_CVPR} employs GAN-tuning \cite{roich2021pivotal} leveraging the face textures UV-maps dataset obtained from real-world scans of AvatarMe++ to extract face diffuse map and lighting information from a single image. Furthermore, FFHQ-UV~\cite{bai2022ffhq} standardizes facial images and creates a high-resolution UV texture dataset using StyleFlow \cite{Abdal2021styleflow} and HIFI3D++ \cite{chai2022realy}, reducing the production cost of the dataset while not retaining the identity of the original in-the-wild images. While recent latent diffusion approaches, such as UV-IDM~\cite{li2024uv} and FreeUV~\cite{yang2025_freeuv}, excel at synthesizing fine-grained facial textures, they typically couple the intrinsic albedo with extrinsic environmental lighting. This inherent entanglement results in baked-in illumination, posing a major obstacle for standard physically-based rendering pipelines when applying novel lighting.

\vsp

\textbf{Text-to-Avatar Generation} Text2face \cite{rowan2023text2face} utilizes StyleGAN2 \cite{karras2020analyzing} to generate in-the-wild images and then leverages CLIP \cite{mohammad2022clip} and Deca \cite{Feng2021deca} to extract image embeddings and 3DMM coefficients \cite{bfm09, blanz2023morphable} from facial images, training an MLP network to establish the mapping between them. During inference, it uses the alignment capability of CLIP between text and image to directly input text embeddings into the MLP to generate 3D faces. Describe3D \cite{wu2023high} manually annotates 1,607 pairs of text and 3D geometric data, training the texture generation model with CLIP loss constraints to achieve inference. Latent3D uses CLIP to iteratively optimize 3DMM coefficients under a single text or image condition to generate and manipulate 3D faces. While these methods to some extent achieve text-to-3D face mapping, they fall short in realistic facial texture generation and do not consider lighting variations. TECA \cite{zhang2023text} and Dreamface \cite{zhang2023dreamface} use the SDS loss \cite{poole2022Dreamfusion} to guide the mapping between text and 3D face, with the former focusing more on hair and accessories generation, and the latter acquiring or utilizing professional equipment to capture real facial texture UV-maps and materials from the real world, training high-fidelity texture generators based on diffusion models. However, such data is costly, scarce, and challenging to ensure model generalization. Due to the bias towards generating a single outcome with the SDS loss, Dreamface undermines the diversity of text-to-3D-face generation.

To address the above issues, we first create a richly annotated text-to-image-to-3D dataset, and then propose PromptAvatar for creating high-fidelity 3D animatable avatars. Our approach integrates an normalized texture UV-map diffusion model and an identity coefficients diffusion model to generate high-quality texture UV-maps and geometric structures for 3D avatars from given prompts. Crafting effective text prompts to generate desired content is a challenging task, and image prompts often contain more information and details than text descriptions \cite{ye2023ip}. So we train a UV-map diffusion model guided by both text and image prompts to cater to diverse user needs.
\section{Dataset Creation}
\begin{figure*}[th]
  \centering
  \includegraphics[width=\linewidth]{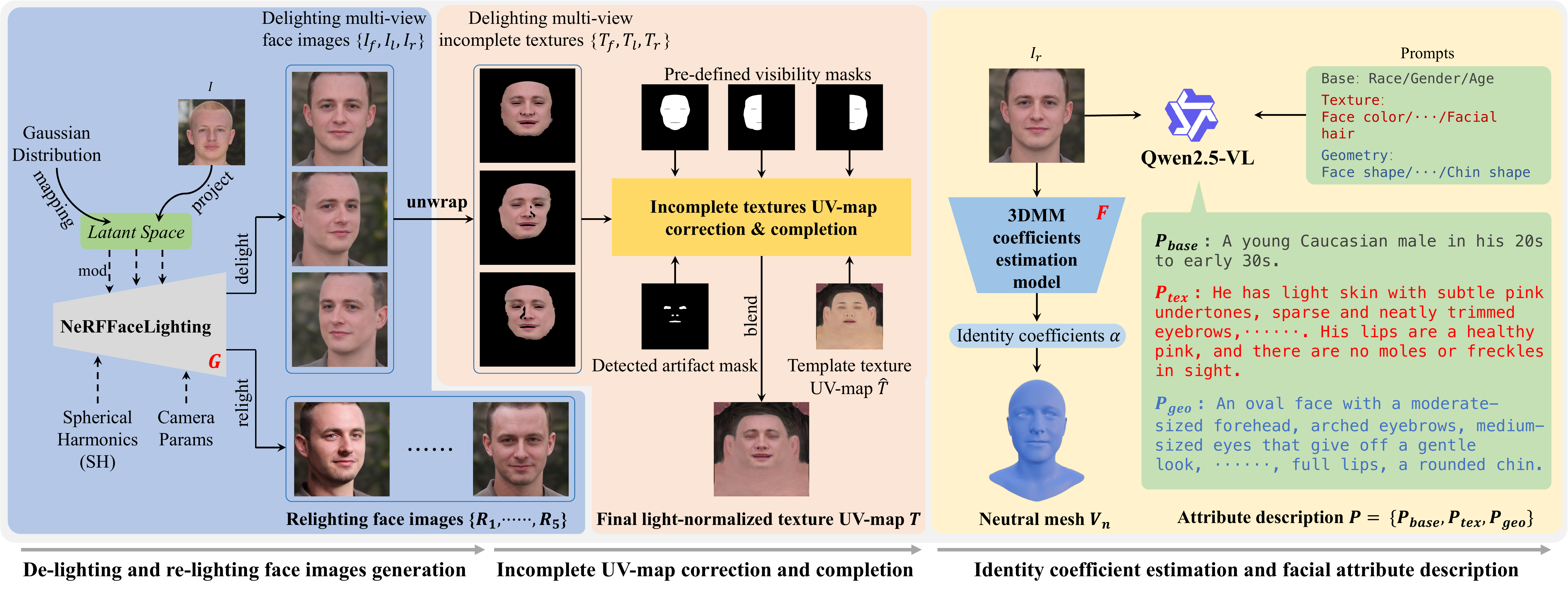}
  \vspace{-5pt}
  \caption{Our dataset creation pipeline consists of three main modules: De-lighting and Re-lighting face image generation, Incomplete UV-map correction and completion, and Identity coefficients estimation and facial attribute description.}
  \label{fig:dataset}
\end{figure*}

We describe the pipeline for constructing our multi-modal dataset, as illustrated in Fig. \ref{fig:dataset}. The process is divided into three primary stages: generating high-quality normalized textures and their corresponding in-the-wild facial images, correcting and completing incomplete UV-maps, and extracting 3D geometric identity coefficients alongside detailed semantic descriptions.

\subsection{De/re-lighting face images generation} 
As shown in Figure \ref{fig:dataset}, our goal is to extract high-quality UV textures from de-lighted multi-view images $\{I_f, I_l, I_r\}$ and simultaneously generate relit facial images $\{R_1, \ldots, R_5\}$ of the same identity under varied lighting and poses. We utilize NeRFFaceLighting \cite{Jiang2023NeRFFaceLighting}, a 3D-aware generative adversarial network (GAN), to maintain superior identity and geometric consistency. Specifically, we project images into a latent space and leverage NeRFFaceLighting to synthesize de-lighted frontal views and side views (at $\pm 35^\circ$ yaw). Finally, we randomly sample spherical harmonics (SH) coefficients and poses from the FFHQ distribution to generate relit images that simulate complex, in-the-wild illumination conditions.

\subsection{Texture correction and completion} 
In this step, we synthesize high-quality facial normalized texture UV-maps from $\{I_f, I_l, I_r\}$. We first "unwrap" the facial images into incomplete UV textures $\{T_f, T_l, T_r\}$ using a pre-trained Deep3D model \cite{chai2022realy} to estimate 3DMM coefficients and calculate vertex coordinates. A facial parsing model is employed to extract masks and filter out non-facial regions. Following the FFHQ-UV \cite{bai2022ffhq} pipeline, we apply linear blending, color correction, and texture completion using predefined visibility masks and template textures $\hat{T}$. This process yields the final high-quality normalized texture UV-map $T$, which is free from environmental lighting and occlusions.

\begin{figure*}[!tbhp]
\begin{center}
\includegraphics[width=0.95\linewidth]{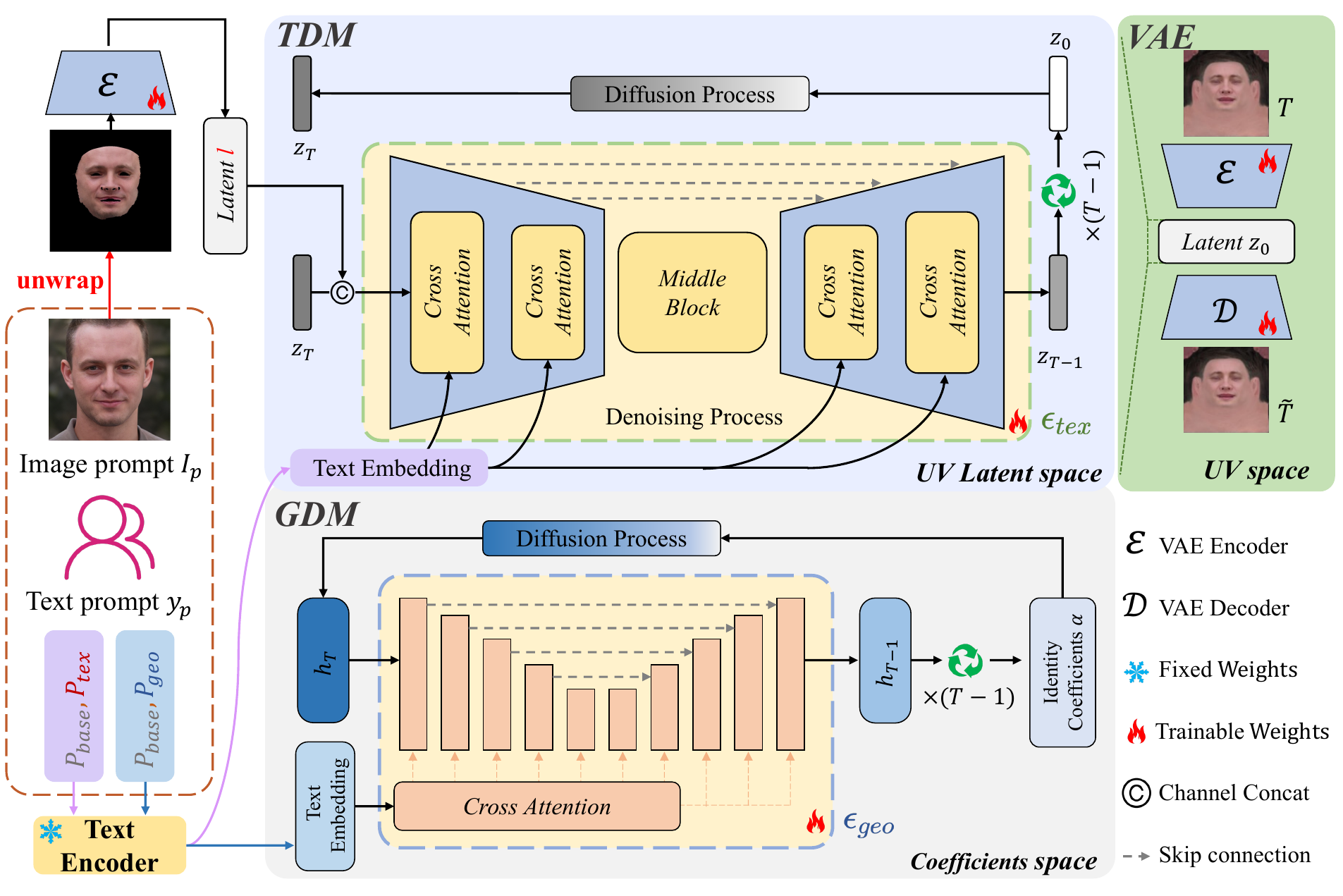}
\end{center}  
  \caption{Architecture of PromptAvatar. The framework comprises a Texture Diffusion Model (TDM), which targets high-quality normalized texture generation, and a Geometry Diffusion Model (GDM) for geometric identity coefficients. Both models are guided by multi-modal prompts embedded via CLIP. For image prompts, incomplete textures are encoded into the latent space to provide localized guidance.}
  \label{fig:method}
\end{figure*}

\begin{table}[htbp]
\centering
\caption{The instruction template used for Qwen2.5-VL-32B-Instruct to generate structured facial attribute descriptions.}
\label{tab:instruction}
\begin{tabular}{|p{0.95\linewidth}|}
\hline
\textbf{Instruction} \\ \hline
Provide a detailed description of this portrait's appearance features within 200 tokens, ignoring hair and expression. Please answer in the following format, for example: \\
\{\{$P_{base}$: A middle-aged male about 50s, Asian yellow human\}, \\
\{$P_{tex}$: Yellow skin with highland red undertones, large pores, thick and long eyebrows, chin and cheeks covered with beard, deep wrinkles around the forehead and eyes, wrinkles between the eyes, freckles distributed on the cheeks, slight eye bags, red lips, deep red nose, dark circles, deep gray beard, black eyebrows, melanin deposits\}, \\
\{$P_{geo}$: rectangular face, overall slender, forehead slightly broad and protruding, willow leaf eyebrows, large almond eyes, relatively narrow eye spacing, high nasal bridge, rounded nose tip, slightly high cheekbones, cherry lips, round but slightly upturned narrow chin, overall carrying a hint of a calm demeanor.\}\} \\
Note that this is very important to my work. Your description should be as detailed and fair as possible. \\ \hline
\end{tabular}
\end{table}

\subsection{Identity and Attribute Extraction} 
To establish multi-modal alignment, we extract identity coefficients and semantic descriptions from the de-lighted frontal image $I_f$. We utilize Deep3D to estimate the facial identity coefficients $\mathbf{\alpha} \in \mathbb{R}^{532}$. Furthermore, to circumvent the costs of manual annotation, we leverage the advanced vision-language reasoning of Qwen2.5-VL-32B-Instruct~\cite{Qwen2.5-VL}. By employing a structured instruction template (as detailed in Table \ref{tab:instruction}), we automatically generate detailed and diverse descriptions $P = \{P_{base}, P_{tex}, P_{geo}\}$. These annotations, averaging 200 tokens, provide fine-grained priors for base demographics, normalized skin textures, and geometric morphology.

\subsection{Data Filtering}
Synthesizing data at a massive scale (initially 500,000 sets) inevitably introduces artifacts such as hair occlusions or specular highlights. We introduce a rigorous dual-stage filtering mechanism to ensure dataset reliability. 
First, we trained an MLP-based aesthetic classifier using CLIP image features, benchmarked against a manually curated set of 2,000 pristine and 10,00 flawed images. This stage eliminates samples with severe occlusions or lighting anomalies. 
Second, we ensure semantic alignment by computing cosine similarity between $I_f$ and its textual description $P$ using CLIP text and image encoders. Only pairs exceeding an empirical threshold are retained. This pipeline distilled our raw data into a highly refined dataset of 100,000 high-fidelity samples.

In summary, we propose a cost-effective and replicable pipeline that integrates NeRFFaceLighting and FFHQ-UV to create a large-scale dataset of 100,000 high-quality normalized textures. Each UV-map is rigorously paired with 3DMM identity coefficients and structured textual descriptions from Qwen2.5-VL-32B-Instruct, forming the foundation for training our PromptAvatar framework.

\section{Dual Diffusion}
\subsection{Preliminaries}
Diffusion models are a type of generative model that first approximate a Gaussian distribution by progressively adding noise to data distributions. Subsequently, a neural network $\epsilon_\theta$ is trained to predict noise $\epsilon$ and perform a denoising process, ultimately generating new data. Diffusion models can be conditioned on other inputs $c$, such as textual descriptions, facilitating the generation of images that are consistent with the specified conditions. This objective is typically formalized through a loss function, expressed as:
\begin{equation}
    \textit{L}=\mathbb{E}_{x_{0}, \boldsymbol{\epsilon} \sim \mathcal{N}(\mathbf{0}, \boldsymbol{I}), c, t}\left\|\epsilon-\epsilon_{\theta}\left(x_{t}, c, t\right)\right\|_{2}^{2},
\end{equation}
where $x_0$ represents the real data, $c$ denotes the input condition, and $\textit{t} \in$ [0, \textit{T}] is the denoising time step.
To better balance image fidelity and sample diversity, classifier guidance \cite{dhariwal2021diffusion} is applied to the score-based models employing gradients from a separately trained classifier. To avoid training an additional classifier, \cite{ho2022classifier} introduces classifier-free guidance, where the diffusion model is trained with and without $c$ randomly during the training process:
\begin{equation}
\hat{\epsilon}_{\theta}(x_t, c, t) = w \epsilon_{\theta}(x_t, c, t) + (1 - w) \epsilon_{\theta}(x_t, \varnothing, t),
  \label{eq:cfg}
\end{equation}
where $w$ is the guidance scale and $\varnothing$ is the embedding of a null prompt.

\subsection{Texture Diffusion Model}
This section introduces our texture diffusion model (TDM), which can generate high-fidelity normalized texture UV-maps using user-provided text prompts and in-the-wild face image prompts. We use the latent diffusion model (LDM) \cite{rombach2022high} architecture to learn the distribution $\textit{p}(T)$ of the UV-maps conditioned on images or textual descriptions. During training, we firstly train the variational autoencoder (VAE) \cite{kingma2013auto}, which contains an encoder $\mathcal{E}$ and a decoder $\mathcal{D}$ in UV-map space. VAE can compress the high-resolution texture UV-map into a latent variable $z_{0}=\mathcal{E}(T) \in \mathbb{R}^{4 \times 64 \times 64}$, which reduces the computational cost and accelerates training.

Subsequently, we fix the weights of the VAE encoder and train the diffusion model (DM) in the latent space based on image and text prompts to establish the mapping relationship between Gaussian noise and texture latent variables. To fully leverage the valuable features provided by an in-the-wild face image prompt, we extract the incomplete UV texture from the image using the ``unwarp'' operation (see Fig. \ref{fig:method}). We encode the incomplete texture UV-map using $\mathcal{E}$ to obtain $l\in \mathbb{R}^{4 \times 64 \times 64}$ and concatenate it with the noisy latent variable $z_t$ at different time steps.
Injecting information about incomplete textures into the diffusion model helps maintain identity consistency between the generated facial texture and the input image. As for the text prompt, we use the textual encoder of CLIP \cite{radford2021learning} to embed it and send the text embedding to the DM's cross-attention layer. Conditioned on both the image prompt $I_p$ and text prompt $y_p$ we train our TDM's denoising model $\epsilon_{tex}$ via the following loss:
\begin{equation}
\small
    \textit{L}_{tex}= \mathbb{E}_{z_0, \epsilon \sim \mathcal{N}(\mathbf{0}, I), I_p, y_p, t}\left\| \epsilon-\epsilon_{tex} \left(z_t, I_p, y_p, t\right) \right\|^{2}_{2},
  \label{eq:uvloss}
\end{equation}
where \textit{t} is the timestep uniformly sampled from $\{1, ..., T \}$ and the latent $\textit{z}_t$ is obtained based on 
\textit{t} and $\textit{z}_0$ \cite{ho2020denoising, rombach2022high}. During the training process, we randomly drop either the image prompt, the text prompt, or both, with a probability of $p_{drop}$, enabling our generation model to adapt to non-prompt, single-prompt, and multi-prompt inputs.
By extracting facial information from text or image prompts, our TDM can generate realistic texture UV-maps with details to aid 3D avatar generation.

\begin{figure}
	\centering
	\includegraphics[width=.85\columnwidth]{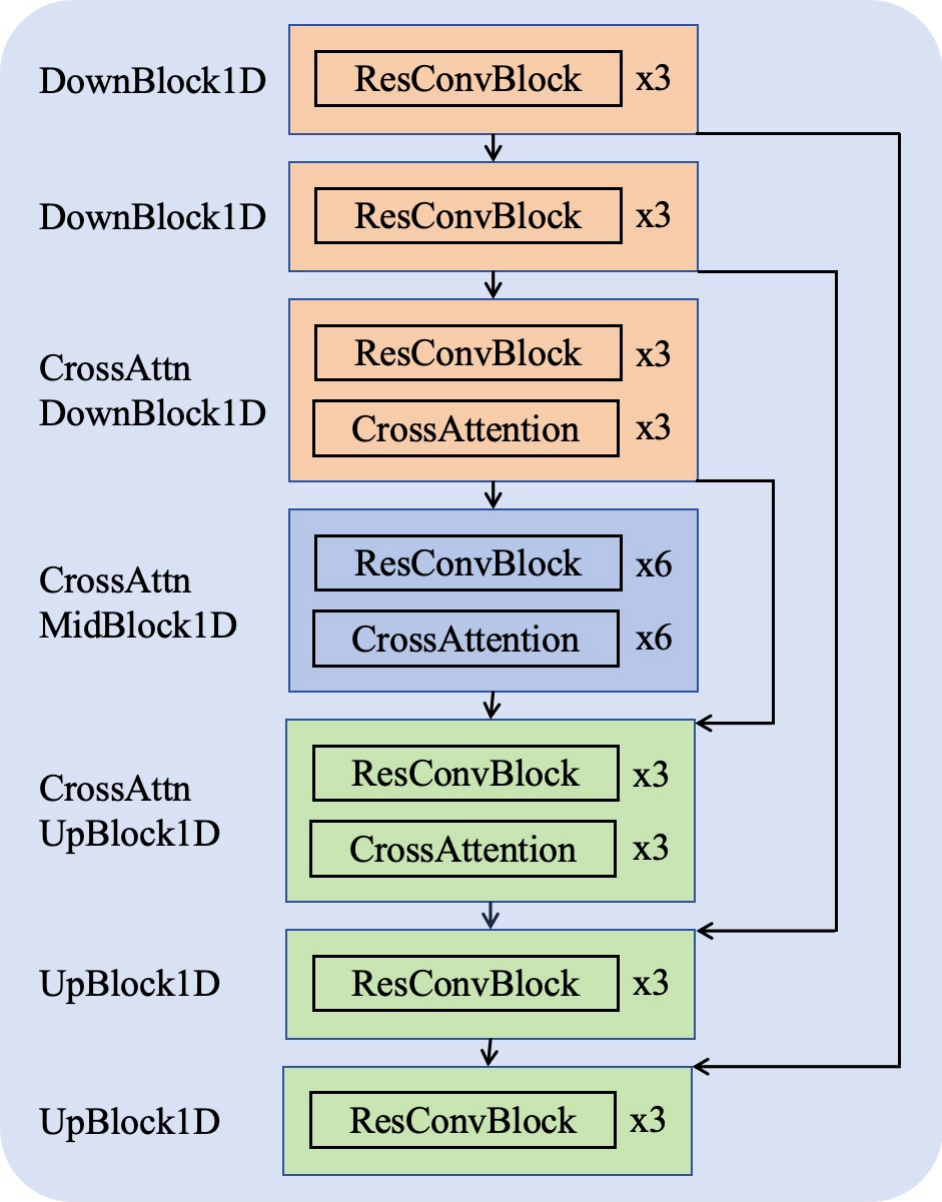}
	\caption{The network architecture of GDM.}
	\label{fig:gdm_arch}
\end{figure}

\subsection{Geometry Diffusion Model}
This section introduces our geometry diffusion model (GDM), which can generate neutral facial geometry guided by a text prompt. Note that in 3DMM, the identity coefficients refer to a set of parameters used to describe the facial shape. By adjusting these coefficients, it is possible to alter features of the model such as facial contour, shape of the eyes, size of the lips, and so on. Identity coefficients play an important role in modeling and controlling the mesh of the face within the 3DMM framework. 

As shown in Figure \ref{fig:method}, we train the GDM in the identity coefficients space. Given the paired data of geometric text prompts $P_{geo}$ and identity coefficients $\alpha$ in the dataset, we utilize CLIP's text encoder to encode $P_{geo}$ into text embedding, enabling GDM to predict the noise applied to the identity coefficients at the time step $t$ under the guidance of the embedding.

Unlike the 2D TDM, the GDM explicitly processes 1D identity coefficients using a custom 1D UNet (ID-UNet) for geometric denoising. As illustrated in Figure \ref{fig:gdm_arch}, the ID-UNet comprises a downsampling encoder, a bottleneck middle block, and an upsampling decoder. The encoder extracts features via 1D downsampling blocks (DownBlock1D), each stacking three residual convolutional blocks (ResConvBlock)[cite: 563, 564, 565]. To inject textual semantic guidance, cross-attention layers (CrossAttention) are explicitly integrated into the third DownBlock1D, the middle block (MidBlock1D, containing six ResConvBlock and six CrossAttention layers), and the first UpBlock1D. The decoder mirrors the encoder, utilizing skip connections to concatenate multi-scale features, and finalizes with purely residual blocks to accurately reconstruct the denoised identity coefficients.

Similar to the training of TDM, GDM is trained via the following loss:
\begin{equation}
    \textit{L}_{geo}= \mathbb{E}_{h_0, \epsilon \sim \mathcal{N}(\mathbf{0}, I), y_p, t}\left\| \epsilon-\epsilon_{geo} \left(h_t, y_p, t\right) \right\|^{2}_{2},
  \label{eq:geoloss}
\end{equation}
where $h_t$ is the identity coefficients with noise added at timestep $t$.

\section{Experiments}
\subsection{Implementation Details} \label{expdetails}
\textbf{Experimental Setup}
For TDM, during training, we use the SD-2.1\footnote{\url{https://huggingface.co/stabilityai/stable-diffusion-2-1}} model to initialize the weights and replicate the weights of the first convolutional layer of the UNet to expand the input dimension to 8. We resize the normalized texture UV-map and the incomplete texture extracted from the image prompt to 512 $\times$ 512 resolution to fit the VAE. We employ the PNDMScheduler \cite{liu2022pseudo} and set the number of time steps $T$ to 1000 during training. We set the sampling steps to 100 during inference, with $w=6$ in Eq. \ref{eq:cfg}. After TDM samples the latent embedding through the UNet based on text or image prompts, the generated texture at 512 $\times$ 512 resolution is fed into $\mathcal{D}$. We then pass this low-resolution texture to a pretrained super-resolution network \cite{wang2021real} to obtain the final 2K-resolution texture UV-map. GDM is trained from scratch, with $w=1.5$ and 200 timesteps during inference.TDM's VAE is finetuned for two days using the Adam optimizer ~\cite{kingma2017adam} with $\beta_1 = 0.5$, $\beta_2 = 0.9$, and learning rate $\eta_1=4.5e-6$. Both TDM's Unet and GDM use AdamW optimizer~\cite{loshchilov2017decoupled} with $\beta_1 = 0.9$, $\beta_2 = 0.999$, weight decay $\mu = 0.01$, and learning rate $\eta_2 =10 \times 10^{-5}$, requiring seven days for the former and two days for the latter. All training is performed on 8 NVIDIA A800-SXM4-80GB GPUs and the inference is on a single GPU. During training, as CLIP's text encoder accepts up to 77 tokens, random combinations and truncations of clauses are done before feeding the text prompt to the encoder to enhance text prompt diversity.

\textbf{Baselines} We first compare our dataset with FFHQ-UV, and then show comparisons against different state-of-the-art approaches for text-to-avatar generation (DreamFace \cite{zhang2023dreamface}, Describe3D \cite{wu2023high}, and Dreamfusion \cite{poole2022Dreamfusion}) and image-to-avatar generation FFHQ-UV and FlameTex \cite{feng2019photometric}). 

\subsection{Evaluation}

\textbf{Dataset}
We measure the superiority of our data compared to FFHQ-UV using two metrics: Identity Similarity and Brightness Symmetry Error (BS Error). The former calculates the identity similarity of multi-view images,  which can reflect the compatibility of the texture UV-maps with their 2D images. We use the face recognition model ArcFace \cite{deng2018arcface} to compute face embeddings for the left-front, left-right, and front-right paired views of multi-view images, normalize them, and calculate the cosine similarity. As indicated in Table \ref{tabsimilarity}, our data set shows a clear advantage compared to FFHQ-UV. 

\begin{table}[thb]  
    \centering  
    \tabcolsep=5pt
    \caption{Quantitative evaluation of Identity Similarity on multi-view images.}  
    \begin{tabular}{@{}c|ccc@{}}   
    \toprule  
    \text{Similarity$\uparrow$} & \text{left-front} & \text{left-right}  & \text{front-right}\\
    \midrule  
    \text{Ours} & \textbf{0.754}  & \textbf{0.801}  & \textbf{0.774} \\
    \text{FFHQ-UV} & 0.634  & 0.544   & 0.632 \\
    \bottomrule  
    \end{tabular}  
    \label{tabsimilarity}  
\end{table}  

BS Error is used to evaluate whether the UV-map has better even illumination as follows \cite{bai2022ffhq}:
\begin{equation}
B S \_E r r o r(T)=\left\|\mathcal{B}_{\alpha}\left(T_{Y}\right)-\mathcal{F}_{h}\left(\mathcal{B}_{\alpha}\left(T_{Y}\right)\right)\right\|_{1},
\end{equation}
where $T_{Y}$ represents the  $Y$  channel of  $T$ in the YUV space; $\mathcal{B}_{\alpha}(\cdot)$  is the Gaussian blurring operation with the kernel size of  $\alpha$  (set to 55);  $\mathcal{F}_{h}(\cdot)$  is the horizontal flip operation. This metric is derived from the observation that a texture map with uneven illumination often exhibits facial shadows, leading to asymmetrical brightness in the UV-map \cite{bai2022ffhq}. As shown in Table \ref{tabbserror}, our texture UV-maps datasets achieve the advantage.

\begin{table}[thb]  
    \centering  
    \tabcolsep=5pt  
    \caption{Quantitative comparison of Brightness Symmetry (BS) Error. Lower values indicate better light-normalization.}  
    \begin{tabular}{@{}c|cc@{}}   
    \toprule  
    \text{Dataset} & \text{Ours} & \text{FFHQ-UV} \\
    \midrule  
    \text{BS Error$\downarrow$} & \textbf{5.098}  & 7.094 \\
    \bottomrule  
    \end{tabular}  
    \label{tabbserror}  
\end{table}

\begin{figure*}[tb]
  \centering
  \includegraphics[width=0.9\linewidth]{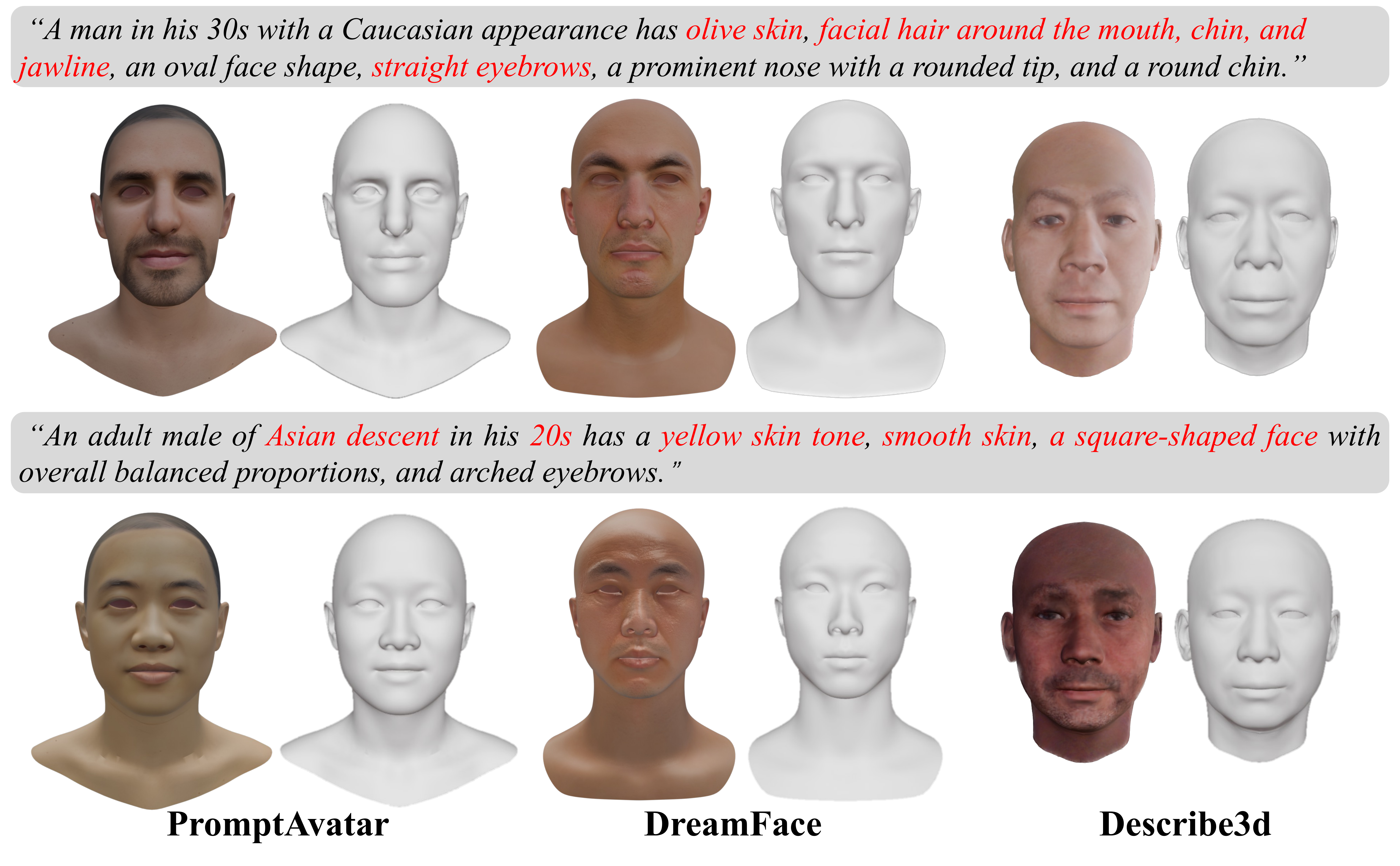}
  \caption{Visual comparison with DreamFace and Describe3D. Our results demonstrate superior alignment with fine-grained text prompts (highlighted in red). For instance, in the first row, observe the distribution of facial hair, eyebrow shape, and chin, and in the second row, note the facial shape, eye bags, and other features.}
  \label{fig:text23D}
\end{figure*}

\vsp

\textbf{Text-To-Avatar} In this section, we conduct qualitative and quantitative comparison experiments on existing text-to-avatar methods such as DreamFace, DreamFusion, and Describe3D. DreamFace and DreamFusion both use the SDS loss to map text prompts to 3D representations, with the former designed explicitly for 3D avatar generation and the latter more adept at general object modeling. Describe3D manually annotates a dataset matching text and 3D information to train a text-to-avatar model and uses the CLIP loss to constrain the rendered results to match the text prompts. Specifically, we select 30 generic facial descriptions as prompts to guide all methods in generating corresponding 3D avatars and rendering them into 2D images. The alignment between the rendered results and the text is measured by calculating the CLIP score. Quantitative results in Table \ref{tab:clipscore} show that our method achieves a better CLIP score and outperforms other methods in inference efficiency. Qualitative results in Figure \ref{fig:text23D} compare two sets of fine-grained text prompts with the results generated from DreamFace and Describe3D. Although Dreamface achieves a more realistic representation in detail due to PBR material considerations, our method can accurately meet various fine-grained prompts, such as facial hair, eyebrow shapes, face shape and so on. This is because the SDS loss sacrifices model diversity to produce smooth results, essentially averaging the information in the text prompts, while our TDM trained directly from text/image to texture can maintain text generation diversity. Additionally, CLIP is trained on general data and is not sufficient to support fine-grained facial descriptions, and thus the results of Describe3D exhibit noticeable flaws.

\begin{table}[t]
    \centering
  \caption{Quantitative comparison with other text-to-avatar methods.}
    \tabcolsep=3pt
    \begin{tabular}{@{}c|cccc@{}} 
    \toprule
    \text{Metric} & \text{DreamFusion}  & \text{Describe3D} & \text{DreamFace} & \text{Ours} \\
    \midrule
    \text{CLIP score$\uparrow$} &20.16 &19.81 & 20.56 & \textbf{21.14} \\
    \text{Infer time(s)$\downarrow$} & 2400 & 15 & 300 & \textbf{10} \\
  \bottomrule
 \end{tabular}
  \label{tab:clipscore}
\end{table}

\begin{figure*}
    \centering
    \includegraphics[width=0.9\linewidth]{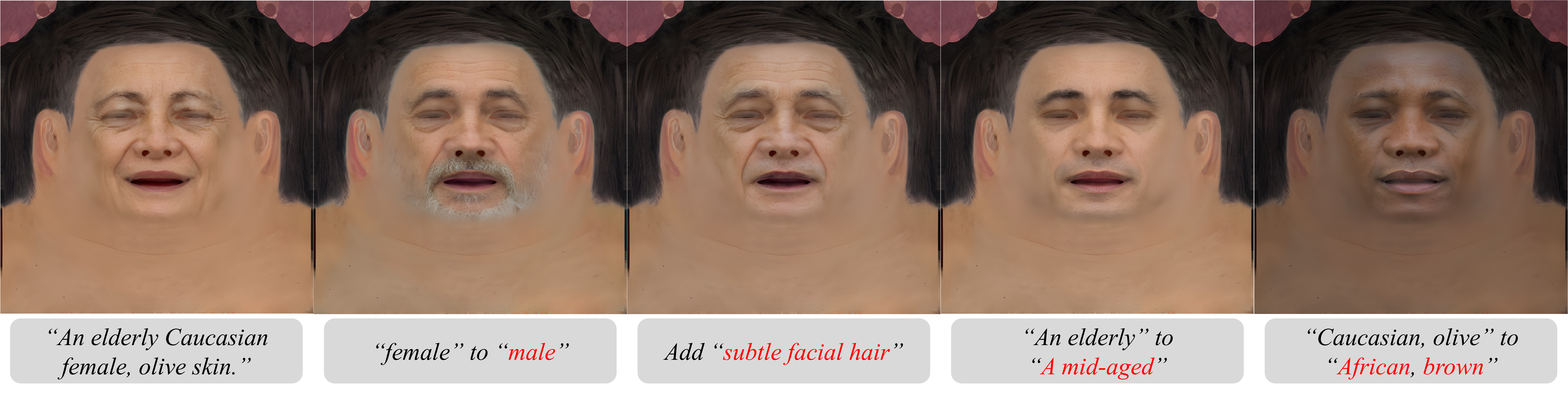}
    \caption{Demonstration of fine-grained facial attribute editing via text prompts. Starting from an initial generated UV-map (left), our Texture Diffusion Model (TDM) accurately manipulates specific semantic features—such as gender, facial hair, age, and ethnicity—by simply modifying the corresponding keywords in the prompts (highlighted in red). The results show that TDM achieves precise localized editing while preserving the unedited features.
    }
    \label{fig:edit}
\end{figure*}

\begin{figure*}[th]
    \centering
    \includegraphics[width=0.95\linewidth]{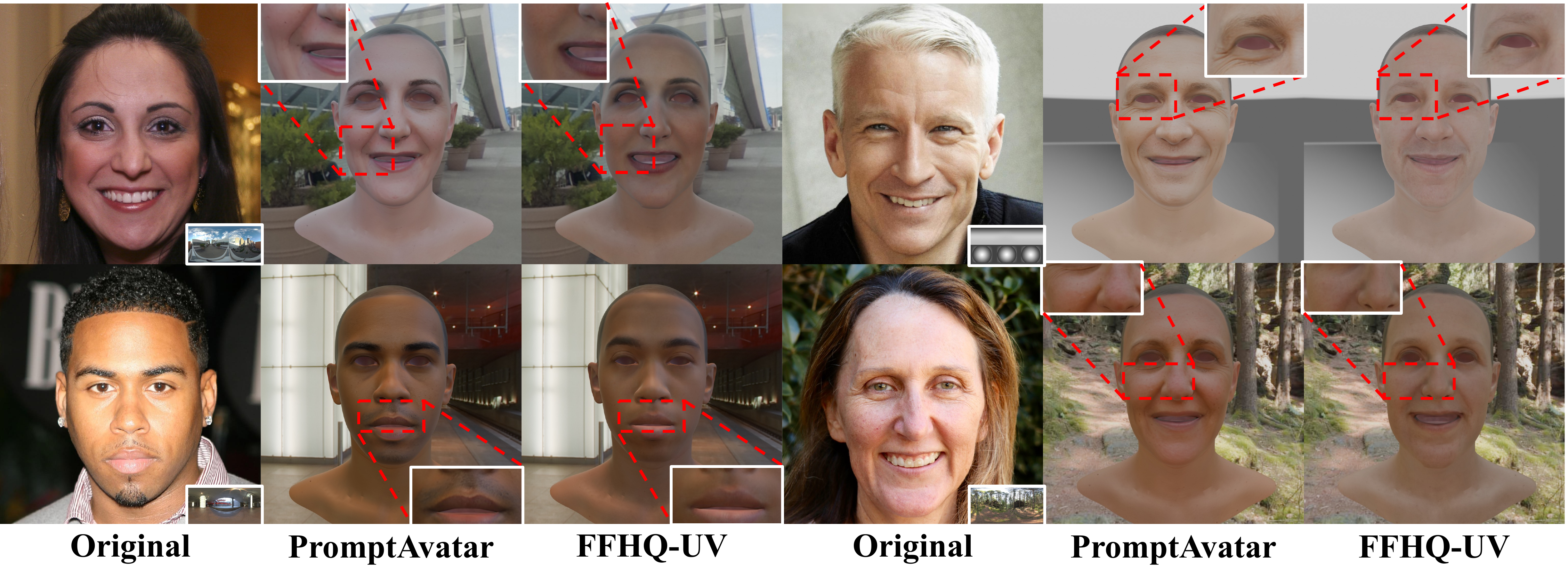}
    \caption{Visual comparison of the image-to-avatar methods. Each group consists of the original image, the re-lit PromptAvatar, and FFHQ-UV. It can be observed that our method is able to preserve more facial details and skin tone of the original image. Please zoom in for better visibility of the details.}
    \label{fig:i2a}
\end{figure*}

\vsp

\textbf{Facial Attribute Editing.} Beyond the feed-forward generation, we further demonstrate the fine-grained facial attribute editing capability of our Texture Diffusion Model (TDM) through sequential text-prompt manipulation. As illustrated in Figure~\ref{fig:edit}, starting from a base avatar, users can progressively refine specific attributes, such as ethnic transitions, the addition of localized facial hair, or the modulation of physiological age and skin tone. These results reveal that TDM successfully disentangles these semantic concepts within its latent representation, facilitating precise, localized modifications while maintaining high identity consistency and preserving unedited regional features.

\begin{table}[t]
    \centering
    \caption{Quantitative experiments of Image-to-Avatar and Ablation studies. CMask-HQ is CelebAMask-HQ.}
    \tabcolsep=3pt
    \begin{tabular}{@{}c|cc|ccc@{}} 
    \toprule
    \text{Similarity$\uparrow$} & \text{FFHQ-UV} & \text{FlameTex} & \text{Ours} & \text{w/emb} & \text{w/both} \\ 
    \midrule
    \text{FFHQ} & 0.354 &0.1589& \textbf{0.370} & 0.284& 0.338 \\
    \text{CMask-HQ} & 0.324 & 0.114 & \textbf{0.334} &0.242 &0.304 \\
    \midrule
    \text{Infer times(s)$\downarrow$} & 80 & 72 &  6 & / & / \\
  \bottomrule
    \end{tabular}
  \label{tabcsim}
\end{table}

\vsp

\textbf{Image-to-Avatar} In this section, we first quantitatively compare PromptAvatar with FFHQ-UV and FlameTex on the FFHQ \cite{karras2019style} and CelebAMask-HQ \cite{CelebAMask-HQ} datasets. FFHQ-UV introduces a large standardized, high-quality texture dataset and trains a StyleGAN2 \cite{karras2020analyzing} model to learn the texture data distribution based on this dataset, fine-tuning weights during actual inference to restore textures from wild images. FlameTex fits the linear texture basis of the FLAME model \cite{FLAME:SiggraphAsia2017} into images through an iterative scheme. For comparison, we render the normalized textures generated by all methods at the resolution of 512$\times$512 and then calculate the identity similarity with the original images. As shown in Table \ref{tabcsim}, our method achieved the best results. Additionally, since FFHQ-UV and FlameTex are iterative fitting methods requiring multiple runs for fine-tuning models or optimizing parameters, while the acceleration methods \cite{Lu2022, song2020denoising} of the diffusion model are well-established, our method has the speed advantage in inference. Qualitative results shown in Figure \ref{fig:i2a} demonstrate that our method outperforms FFHQ-UV in detail restoration and skin tone preservation. For example, in the first row, see the eye bags of the female, crow's feet, and smile lines of the male, and in the second row, see the beard of the male, freckles of the female, and so on. The difference between the two lies in the fact that GANs tend to produce smooth results, while the linear model used by FlameTex is insufficient to express high-frequency facial details.

\vsp

\textbf{Ablation Studies} To maximize the utilization of facial information contained in image prompts, we compare the impact of different forms of conditional guidance on the performance of the texture diffusion model. As shown in Table \ref{tabcsim}, ``w/emb'' denotes the process of extracting identity embeddings from in-the-wild facial images through the pre-trained face recognition network ArcFace \cite{deng2018arcface}, feeding these embeddings into a trainable projection network to obtain image tokens, and then guiding model training through a set of cross-attention layers parallel to those traversed by the text prompts similar to InstantID \cite{wang2024instantid}, specifically designed to encode facial identity embeddings into the feature space. ``w/both'' represents a hybrid approach combining our method and the ``w/emb'' method to extract image information for guiding model training to restore high-quality textures. It is important to note that the weights of the UNet in all three training methods need to be fine-tuned to adapt to the latent space distribution of texture UV-maps. Quantitative results indicate that our method achieves the highest identity similarity in the image-to-avatar experiment. We attribute this to the fact that the embeddings extracted by the pre-trained face recognition network overlook the facial detail features. 

\begin{figure*}[t]
    \centering
    \includegraphics[width=0.9\linewidth]{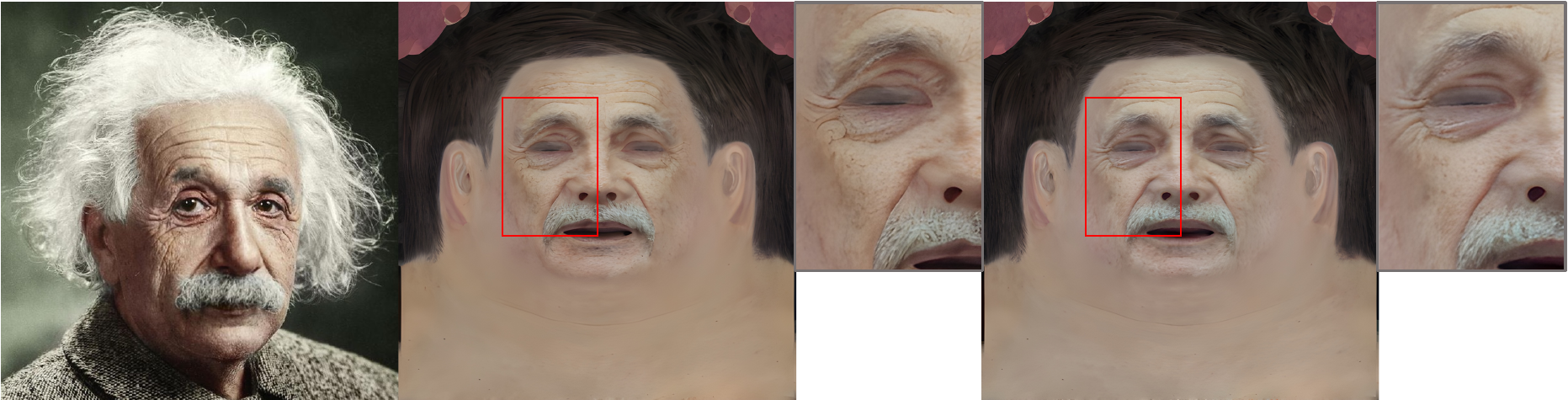}
    \caption{Ablation study on VAE fine-tuning. Left: Original in-the-wild image. Middle: Texture UV-map generated using the frozen pre-trained SD-2.1 VAE, which results in overly smoothed high-frequency details. Right: Texture generated using our fine-tuned VAE. The fine-tuned VAE successfully adapts to the UV-map data distribution, effectively recovering realistic facial features such as crow's feet and wrinkles (highlighted in the red boxes).}
    \label{fig:vae}
\end{figure*}

We also perform ablation experiments to evaluate fine-tuning the VAE within the TDM pipeline. The pre-trained SD-2.1 VAE was optimized for natural image distributions, which differ from the unwrapped topology of UV-maps. As shown in Figure~\ref{fig:vae}, freezing the pre-trained VAE results in overly smoothed textures lacking high-frequency realism. Conversely, our fine-tuned VAE adapts to the distinct latent distribution of texture UV-maps, enabling more realistic facial details, such as crow's feet and under-eye bags.

\section{Discussion} \label{discussion}
The proposed PromptAvatar framework shifts the paradigm of 3D avatar generation from prior iterative optimization~\cite{bai2022ffhq, zhang2023dreamface, rowan2023text2face} to a more efficient direct synthesis approach. By leveraging a constructed large-scale four-modality dataset, our dual diffusion models for geometry and texture successfully bridge the gap between diverse multi-modal prompts and high-fidelity 3D representations, without requiring time-consuming SDS or CLIP-based fine-tuning. Moreover, our architecture can recover high-frequency facial details such as pores and wrinkles within 10 seconds. This efficiency significantly lowers the barrier to generating personalized digital assets compared to existing iterative fitting or scan-based approaches.

Despite these advancements, several limitations remain that suggest directions for future research. First, although our dataset is large-scale and highly diverse, it is synthesized based on the distribution of the FFHQ dataset~\cite{karras2019style} and may still inherit certain underlying biases in texture attributes. Second, we observed that the 3DMM~\cite{bai2022ffhq} identity coefficients used in this study can be sensitive to facial expressions, which might introduce slight geometric distortions in the neutral mesh. Transitioning to more robust parametric models or developing expression-invariant geometry encoders could further enhance structural accuracy. Finally, the current framework primarily focuses on light-normalized texture and geometry. Future work will explore the generation of additional material properties, such as roughness, metallicness, and normal maps, as well as the integration of end-to-end neural rendering to further refine the alignment between multi-modal prompts and 3D outputs.

\vsp

\section{Conclusion} \label{conclusion}
In this paper, we present a large-scale dataset with four modalities linking text and/or images to 3D avatars, and introduce PromptAvatar, a method that quickly generates accurate 3D facial UV maps and geometry from text and/or image prompts. Extensive experiments demonstrate PromptAvatar's superior efficiency, diversity, and quality compared to existing methods. By releasing the dataset publicly, we aim to advance research in text-image-prompted 3D avatar generation.

{
    \small
    \bibliographystyle{unsrt}
    \bibliography{main}
}

\appendix
\clearpage
\section{Apendix}

\subsection{Text Prompts for Avatar Generation}
\label{sec:appendix_prompts}

To quantitatively and comprehensively evaluate the generative capabilities and diversity of our proposed PromptAvatar in Table~\ref{tab:clipscore}, we constructed a diverse set of text prompts. Table~\ref{tab:prompts_coarse} presents the coarse-grained descriptions, while Table~\ref{tab:prompts_fine} details the fine-grained, attribute-specific descriptions.

\begin{table*}[h]
\centering
\renewcommand{\arraystretch}{1.3} 
\begin{tabular}{p{0.95\linewidth}} 
\toprule
\multicolumn{1}{c}{\textbf{Part I: Coarse-grained Prompts}} \\
\midrule
55-year-old Caucasian male, rosy cheeks, noticeably thin face with slightly sunken cheeks. \\
African child, round face, small eyes, thick lips, short curly hair. \\
Elderly Caucasian male, elegant appearance, big eyes, long pointed nose with visible wrinkles, wide lips. \\
30-year-old Middle Eastern male, prominent nose, thick eyebrows, medium beard, calm and regal facial expression. \\
15-year-old female, cute appearance, round face, bright eyes, straight black hair. \\
25-year-old female, captivating smile, elegant facial features, wavy blonde hair. \\
20-year-old European Caucasian female, defined jawline, thick lips, heavy black eyeliner. \\
Young female, cute and delicate facial features, youthful appearance, soft lighting. \\
Female in her early 20s, heart-shaped face, small nose, full lips, vibrant and youthful look. \\
27-year-old male of African descent, unusually light skin tone, striking facial symmetry. \\
Young Asian female, delicate facial features, big eyes, healthy rosy complexion, gentle smile. \\
Male, round face with a surprisingly sharp jawline, small nose, short hair. \\
Middle-aged Western male, big dark eyes with double eyelids, high nose bridge, dense square black beard. \\
70-year-old male, big round nose, balding head, tired eyes. \\
Male, striking prominent cheekbones, deep-set charming eyes, clean-shaven. \\
Male, prominent cheekbones, sharp pointed chin, small nose, intense gaze. \\
Male, rugged appearance, spiky black hair, strong square jaw, five o'clock shadow. \\
60-year-old male, salt and pepper hair, square receding hairline, stern and serious expression. \\
Elderly male, gentle and smiling eyes, neat gray hair, wearing reading glasses. \\
90-year-old male, heavy-set face, chubby cheeks, unkempt white beard, unfriendly glare. \\
\bottomrule 
\end{tabular}
\caption{Extended list of text prompts (Part I). These coarse-grained descriptions focus on general demographic attributes and basic facial features, explicitly omitting non-facial body descriptions.}
\label{tab:prompts_coarse}
\end{table*}

\begin{table*}[!htbp]
\caption{Extended list of text prompts (Part II). These fine-grained descriptions incorporate diverse and challenging facial details such as pores, wrinkles, acne, and pigmentation.}
\centering
\renewcommand{\arraystretch}{1.3} 
\begin{tabular}{p{0.95\linewidth}} 
\toprule 
\multicolumn{1}{c}{\textbf{Part II: Fine-grained Prompts}} \\
\midrule
A Caucasian male in his 30s with medium-tone skin and neutral undertones has some visible pores, very light stubble on the chin and upper lip, a few faint smile lines around the mouth, minimal creases under the eyes, and slightly noticeable frown lines on the forehead. \\
A mixed man in his 40s with a personable appearance has fair skin and rosy cheeks. His complexion is clear with subtle lines under his eyes and faint laugh lines around his mouth. \\
A young adult Caucasian male in his 30s with a professional appearance has fair skin with slight pink undertones, small pores, and a clean-shaven look. \\
A 30-year-old light-skinned adult male with olive skin tone and faint pink undertones has small pores and slight nasolabial fold visibility. \\
A Caucasian male in his 30s has light skin with subtle pink undertones and visible pores. He has a small number of faint freckles on the cheeks and minimal signs of wrinkles, mainly in the nasolabial fold area. \\
A middle-aged Asian male in his 50s with a professional demeanor. His skin features relatively small pores, faint fine lines, and wrinkles mainly on the forehead and around the eyes, with some visible around the mouth. \\
A youthful individual with a fair human complexion in their early teens possesses smooth fair skin with peachy undertones, fine pores, delicate and sparse eyebrows. \\
A young adult male with light brown skin appears to be in his late 20s to early 30s, displaying an overall smooth complexion with subtle uneven skin tones and minimal pores. \\
A Middle-Eastern woman in her early 30s has fair skin with subtle pink undertones, moderately sized pores, sparse and fine eyebrows, faint lines on the forehead. \\
A young Caucasian female in her 20s with light olive skin featuring subtle pink undertones. She has fine pores and neatly shaped eyebrows with no stray hairs. \\
An elderly African woman in her 70s with rich dark brown skin and cool undertones. Her face features deep wrinkles across the forehead and around the eyes, prominent cheekbones, and slight hyperpigmentation on the cheeks. \\
A South Asian male in his late 20s with warm golden undertones. He has a thick, well-groomed dark beard, dense eyebrows, and faint dark circles under his expressive eyes, with a generally smooth skin texture. \\
A Hispanic female in her 40s with a medium olive complexion. Her face displays subtle sun spots near the temples, thick naturally arched eyebrows, faint crow's feet, and she wears bold red lipstick. \\
An elderly East Asian woman in her 80s with thin, slightly translucent skin. Her face is characterized by deep nasolabial folds, prominent age spots on the cheekbones, hooded eyelids, and sparse, silvery-white eyebrows. \\
An Indigenous male in his 50s with weathered, copper-toned skin. He has deep, prominent forehead lines, strong and high cheekbones, a faint scar on his left cheek, and a rugged, unshaven jawline. \\
A young adult of African descent in their 30s with vitiligo. Their face features striking, symmetrical patches of depigmented skin around the eyes and mouth, contrasting beautifully with their natural deep espresso skin tone, paired with a neatly trimmed goatee. \\
A Caucasian teenage girl with very fair skin and prominent reddish undertones. Her face shows scattered freckles across the bridge of her nose and some active acne blemishes along the jawline, with thick, unplucked natural eyebrows. \\
A mature Mediterranean man in his 60s with deeply tanned, leathery skin. He has prominent laugh lines, bushy salt-and-pepper eyebrows, a thick gray mustache, and noticeable pores on his nose. \\
A young woman with albinism in her early 20s. She has extremely pale, almost translucent skin with very subtle pinkish undertones, framed by pale blonde eyelashes and nearly invisible eyebrows, with a flawless, poreless texture. \\
A Southeast Asian male in his 40s with a warm tan complexion. He features distinct acne scarring on his lower cheeks, a broad nose, slightly oily skin texture, and a faint mustache. \\
\bottomrule
\end{tabular}
\label{tab:prompts_fine}
\end{table*}

\begin{figure*}[th]
    \centering
    \includegraphics[width=1\linewidth]{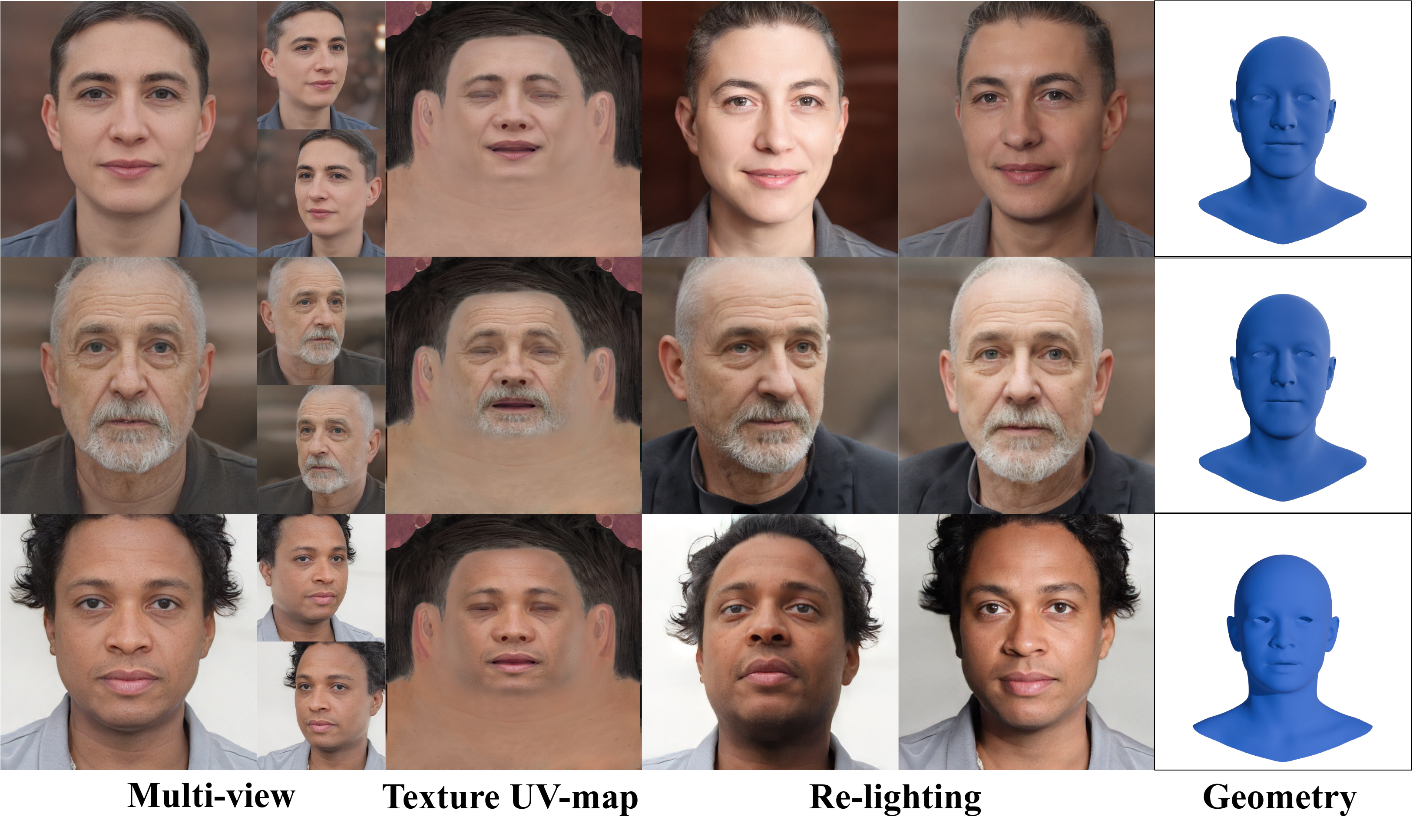}
    \caption{\textbf{Visual examples of the multi-modal facial dataset.} Each row displays a unique identity from our collection. From left to right: multi-view source images, unwrapped high-resolution texture maps, facial renderings under novel lighting, and the corresponding 3D geometry. }    
    \label{supp:dataset}
\end{figure*}

\subsection{Visualization of Dataset}
We visualize a subset of the data in Figure \ref{supp:dataset}, where each row contains multi-view images, high-quality texture maps, re-lighting facial images, and geometric structures.

\begin{figure*}[thb]
  \centering
  \includegraphics[width=0.9\linewidth]{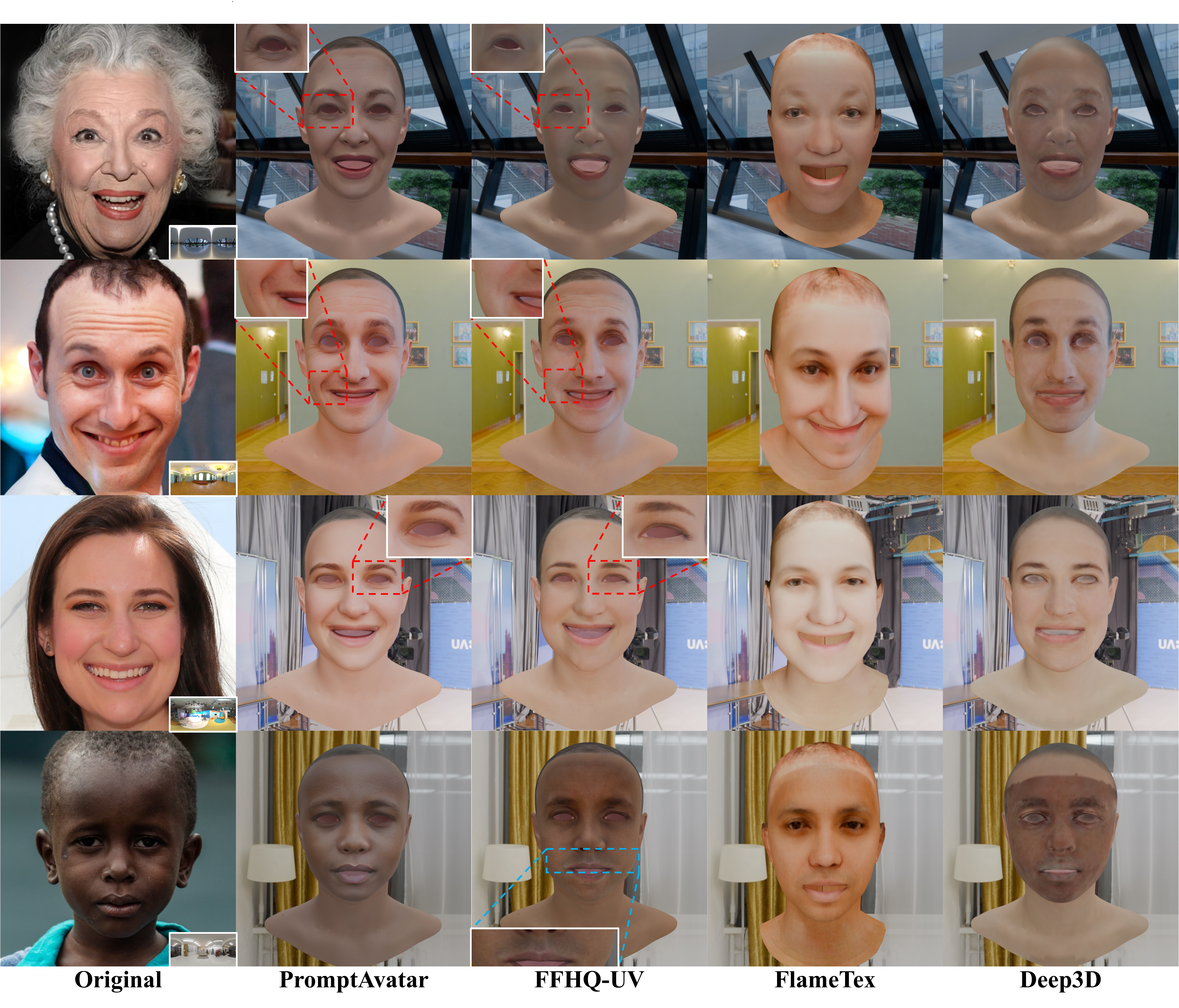}
  \caption{More visual comparison of the image-to-avatar methods. The \red{red dashed boxes} indicate that our method exhibits better details than FFHQ-UV, while the \blue{blue dashed boxes} indicate flaws in FFHQ-UV. The bottom-right corner of each image in the first column is the environment map used for rendering. The FlameTex and Deep3D results use textures directly estimated by each method for visualization.}
  \label{fig:supp_comparison}
\end{figure*}

\subsection{Qualitative Comparisons on Diverse Identities}
To further evaluate the generalization and robustness of our proposed method, we present additional visual comparisons with several state-of-the-art methods supporting normalized texture generation (e.g., FFHQ-UV, FlameTex, and Deep3D) in Figure \ref{fig:supp_comparison}. 
Unlike existing approaches that often produce over-smoothed albedo maps or retain residual environmental lighting, our method effectively decouples illumination from identity-specific features. 
As highlighted in the zoomed-in regions, our approach demonstrates superior capability in recovering fine-grained, shading-free textures (e.g., pores and wrinkles) even from "in-the-wild" images with complex expressions and diverse ethnicities.

\end{document}